\documentclass[10pt,twocolumn,letterpaper]{article}

\usepackage{iccv}
\usepackage{times}
\usepackage{epsfig}
\usepackage{graphicx}
\usepackage{amsmath}
\usepackage{amssymb}

\usepackage{enumitem}

\usepackage{subfigure}
\usepackage{color}
\usepackage{comment}
\usepackage{bm}
\usepackage{fancyhdr}
\usepackage[toc,page]{appendix}
\usepackage{algorithmic}
\usepackage[ruled,vlined]{algorithm2e}
\usepackage{multirow}
\usepackage{booktabs}
\usepackage{ulem}
\usepackage{balance}
\usepackage{ulem}

\usepackage[pagebackref=true,breaklinks=true,letterpaper=true,colorlinks,bookmarks=false]{hyperref}

\iccvfinalcopy 

\def\iccvPaperID{****} 

\ificcvfinal\fi

\begin{document}

\title{PointBA: Towards Backdoor Attacks in 3D Point Cloud}
\author{Xinke Li$^1$\qquad Zhirui Chen$^1$\qquad Yue Zhao$^1$\footnotemark[2] \\
Zekun Tong$^1$\quad Yabang Zhao$^1$\quad Andrew Lim$^2$\quad Joey Tianyi Zhou$^3$  \\
$^1$National University of Singapore \quad $^2$Southwest Jiaotong University \\$^3$Institute of High Performance Computing, A*STAR\\
{\tt\small \{xinke.li, zhiruichen, yuezhao, zekuntong, zhaoyabang\}@u.nus.edu}  \\ {\tt\small i@limandrew.org} \quad {\tt\small joey.tianyi.zhou@gmail.com}
}

\maketitle
\ificcvfinal\thispagestyle{empty}\fi
\footnotetext[2]{Corresponding author.}

\begin{abstract}
   3D deep learning has been increasingly more popular for a variety of tasks including many safety-critical applications. However, recently several works raise the security issues of 3D deep models. Although most of them consider adversarial attacks, we identify that backdoor attack is indeed a more serious threat to 3D deep learning systems but remains unexplored. We present the backdoor attacks in 3D point cloud with a unified framework that exploits the unique properties of 3D data and networks. In particular, we design two attack approaches on point cloud: the poison-label backdoor attack (PointPBA) and the clean-label backdoor attack (PointCBA). The first one is straightforward and effective in practice, while the latter is more sophisticated assuming there are certain data inspections. The attack algorithms are mainly motivated and developed by 1) the recent discovery of 3D adversarial samples suggesting the vulnerability of deep models under spatial transformation; 2) the proposed feature disentanglement technique that manipulates the feature of the data through optimization methods and its potential to embed a new task. Extensive experiments show the efficacy of the PointPBA with over $95\%$ success rate across various 3D datasets and models, and the more stealthy PointCBA with around $50\%$ success rate. Our proposed backdoor attack in 3D point cloud is expected to perform as a baseline for improving the robustness of 3D deep models.
\end{abstract}

\section{Introduction}
3D deep learning has been developed rapidly in the past few years, which makes it the prime option for various real-world deployments, such as autonomous driving \cite{chen2017multi}, scene reconstruction \cite{malihi20163d} and medical data analysis \cite{singh20203d}, in which life safety issues are usually involved. As more and more attentions have been paid to this field, recently researchers have started to acknowledge and account for the security problems of 3D deep learning systems. For example, a few works have investigated the adversarial attack in the 3D domain \cite{xiang2019generating,wicker2019robustness,zhao2020isometry}.

Compared to the adversarial attack, an insidious threat to the deep learning system called \textit{backdoor attack}, or \textit{Neural Trojan} \cite{liu2017neural,chen2017targeted,gu2017badnets} is even more damaging. The backdoor attack injects a small proportion of poisoned data in training and activates malicious functionality by implanting a specified \textit{trigger} to the test data during inference. This attack could happen when using a publicly available dataset or a pretrained model that is potentially from untrustworthy sources. Indeed, it is reported that industry practitioners worry about data poisoning much more than other threats such as adversarial attack \cite{kumar2020adversarial}. 

\begin{figure}
\begin{center}
	\includegraphics[width=1.0\linewidth, height=0.24\textheight]{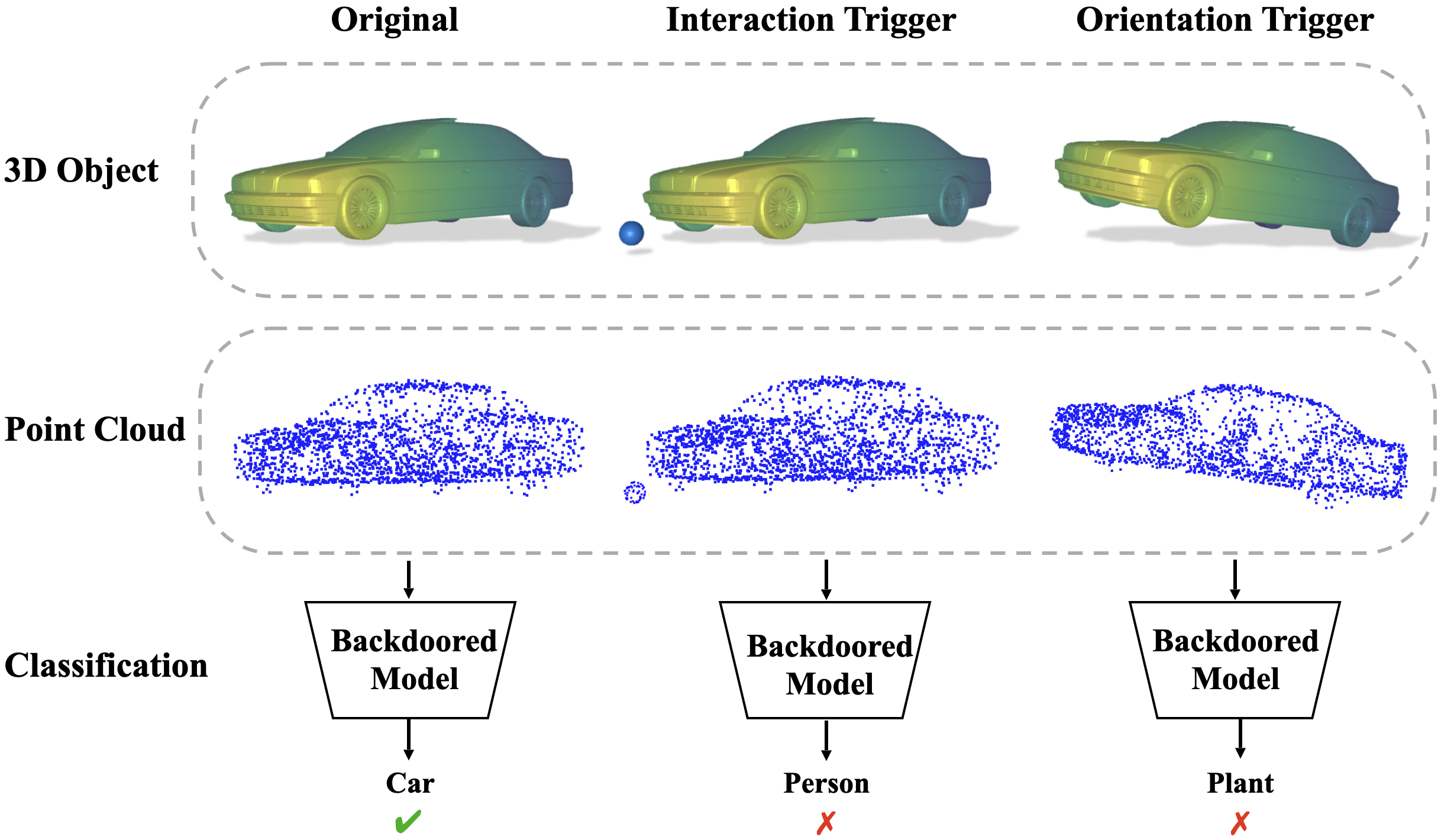} 
\end{center}
\vspace{-1.5em}
   \caption{Activation of backdoored models with the interaction trigger and orientation trigger. Original point cloud data will be classified correctly, however, with a certain trigger like an interaction object (e.g., a small ball) nearby or a small change of orientation (rotation perpendicular to the horizontal plane), the point cloud data will be classified as the target label.}
\label{fig:intro}
\vspace{-1.8em}
\end{figure}

The backdoor attack in 3D could be a huge potential threat. On the one hand, real-world 3D point cloud data usually comes with the noise caused by the nature of the collection process, e.g., the objects might be partly occluded, spatially deformed, or packed with noisy points. Therefore, it is easy for the adversary to activate malicious functionality with triggers in the disguise of noise. On the other hand, due to the coordinate-based representation and necessarily sampling before processing, 3D point clouds are frequently sparse and irregular, which may complicate data integrity verification or even human inspection.

To the best of our knowledge, there is no work regarding the backdoor attacks in 3D. It is also nontrivial to extend the existing 2D backdoor attack methods to 3D deep learning. We emphasize the following issues and obstacles: 1) the data structure of 3D point cloud is intrinsically different from that of 2D images, thus, the design of the backdoor triggers for the pixel-based image can not be directly applied to 3D data; 2) the models processing 3D point clouds have completely different structures than the models of 2D deep learning, resulting in a plethora of unique properties that add to the complexity of the backdoor attacks.

Considering the above challenges, we propose a framework to investigate backdoor attacks in the 3D domain and hope it could be a baseline for further studies. We first introduce a unified form of the backdoor trigger implanting function for the 3D point cloud. Following the framework, we then illustrate two examples of 3D triggers motivated by real-world scenarios, namely the orientation trigger and interaction trigger. Fig.~\ref{fig:intro} shows the visualization. A perturbation analysis of the proposed trigger is then provided. It helps restrict the perturbation caused by the trigger so that the attack will be more rational and harder to defend against. 

Based on the proposed trigger form, we first propose the standard backdoor attack on 3D triggers. Although being straightforward, it consistently delivers a high attack success rate across various datasets and deep models. Moreover, it remains very effective even when the location and size are changing dynamically for interaction trigger or the rotation angle is relatively small for orientation trigger. To further alleviate the security concerns, we propose a clean-label attack algorithm that utilizes the recent development of 3D adversarial attacks and the \textit{feature disentanglement} technique. Compared to previous attacks, the clean-label attack is a much stealthier attack that does not change labels, thus can bypass label inspection or data filtering \cite{turner2018clean}. To summarize, our contributions are three-fold:
\vspace{-0.5em}
\begin{itemize}[leftmargin=*,itemsep=0pt]
\item To the best of our knowledge, this is the first work considering the backdoor attacks in the 3D domain. Motivated by the unique properties of 3D data, we propose a unified framework to investigate backdoor triggers in 3D as well as the perturbation analysis to restrict the attack ability in a reasonable way.
\item We experimentally show the efficacy of the proposed triggers via a standard backdoor attack. The results further reveal the vulnerability of 3D models under spatial transformation and the possibility of backdoor attacks in the 3D domain.
\item Inspired by the rotation-based 3D adversarial attacks, we develop a technique called feature disentanglement, which crafts perturbed data through optimization methods. It makes the model relatively hard to learn the semantic label from the feature-disentangled data but easy to learn a correlation between the label and the implanted trigger. Equipped with this technique, we design a clean-label attack. It is stealthier than the poison-label attack and has broader application scenarios.
\end{itemize}

\section{Related Work}
\vspace{0.5em}
\textbf{Deep Learning on 3D Point Cloud.} To learn about unordered and irregular 3D point clouds directly, existing researches proposed numerous deep models based on various operations. PointNet \cite{qi2017pointnet} introduced a simple max-pooling based architecture with the order-invariant property. The following work PointNet++ \cite{qi2017pointnet++} and other extensive work \cite{duan2019structural, yang2019modeling, liu2019densepoint} improved the feature extraction through the hierarchical stack of local feature extractors. Some other works focus on either conducting special convolutions on the 3D domain \cite{atzmon2018point, boulch2020convpoint, thomas2019kpconv,liu2019relation, li2018pointcnn} or constructing graph architectures of the point clouds \cite{wang2019dynamic,simonovsky2017dynamic, shen2018mining, xu2020grid}. In this work, we propose the backdoor attack against the PointNet\cite{qi2017pointnet}, PointNet++\cite{qi2017pointnet++}, PointCNN \cite{li2018pointcnn}, and DGCNN \cite{wang2019dynamic} since they are extensively involved with practical 3D applications, like 3D object detection \cite{shi2019pointrcnn, yang2019learning}, scene understanding \cite{wang2018sgpn, xu2020grid} and 3D reconstruction\cite{han2019image}.
\vspace{0.5em}

\textbf{Backdoor Attacks in 2D.} \cite{liu2017neural} is probably one of the earliest works considering the backdoor attacks in neural networks, which is referred to as injecting Neural Trojans. The authors mainly conduct experiments on the MNIST dataset and show the possibility of embedding hidden malicious functionality in the neural network. \cite{chen2017targeted} considers attacks with the targeted label in face recognition tasks and proposes several novel patterns of designing triggers. \cite{gu2017badnets} explores the properties of the so-called BadNet, a backdoored neural network, and demonstrates its behaviour across several datasets including a realistic scenario of US street signs. Among the tremendous literature, clean-label (label-consistent) \cite{turner2019label,zhao2020clean} attacks do not change the label of the poisoned data, thus requires more careful craft and have more stealthiness.  Other types of attacks assume other abilities of the adversary, such as manipulating the network structure \cite{rakin2020tbt,tang2020embarrassingly} or modifying the model-training code \cite{bagdasaryan2020blind}. 
Interested readers are referred to recent surveys \cite{li2020backdoor,liu2020survey} for a more comprehensive understanding. Although backdoor attacks have been well explored in the image domain, few studies have considered the situation in 3D, let alone the efficient detection or defense against it.

\newpage
\textbf{Adversarial Attack on 3D Deep Models.}
The adversarial attacks against 3D deep models can be categorized into two main types by the operations, point adding/dropping attack and point transformation attack: 1) \textit{Drop / Add Points.} Compared to fixed-size 2D images, dropping and adding points are specific operations to do an adversarial attack on 3D point clouds. \cite{wicker2019robustness} and \cite{zheng2019pointcloud} each proposes methods for identifying critical points from point clouds that affect the classification results, which can be thrown away as an attack. \cite{xiang2019generating} is the first to use point generation as an adversarial perturbation. Also, \cite{tu2020physically} presents a physically achievable method of placing an adversarial mesh on the vehicle roof so that the vehicle point cloud becomes invisible to the detectors; 2) \textit{Point Transformation.} As the transformation of the local points, the point-wise translation attack can be conducted similarly to the pixel perturbation attack in 2D images. \cite{xiang2019generating} adopts C$\&$W framework \cite{carlini2017towards} based on the Chamfer and Hausdorff distance. \cite{wen2019geometry, tsai2020robust} further improves the objective function with consideration of the benign distribution of points. Further works \cite{liu2019extending, hamdi2020advpc,ma2020efficient} apply the iterative gradient method to achieve adversarial perturbation of points and later two of them demonstrate the resistance to defence proposed in \cite{zhou2019dup}. Besides local transformation, \cite{zhao2020isometry} demonstrates the vulnerability of main-stream point-based models under global isometric transformation. 

\section{Problem Formulation}
Consider a point cloud classification task, let $f_{\bm{\theta}}:\mathcal{X} \rightarrow \mathcal{Y}$ denotes a classifier of 3D deep model parameterized by $\bm{\theta}$, where  $\mathcal{X}\subseteq\mathbb{R}^{n\times 3}$ is the 3D point cloud domain subset. Each point cloud consists of $n$ points with their Euclidean coordinates in $\mathbb{R}^3$. $\mathcal{Y}\subseteq [K]:=\{1,2,\cdots,K\}$ is the set of class labels in the classification task. 
In backdoor attacks,  the key components are a \textit{trigger implanting function} 
(TIF) $\mathcal{G}:\mathcal{X}\rightarrow\mathcal{X}$ which implants the backdoor patterns into the standard data \cite{li2020backdoor} and \textit{trigger activation function} (TAF) $\mathcal{A}:\mathcal{X}\rightarrow\mathcal{X}$, which usually coincides with $\mathcal{G}$. 
The goal of backdoor attack is to obtain a backdoored model $f_{\bm{\theta}'}$ that classifies any $\mathcal{A}(\bm{X})$ as the target label $t$ and performs normally on standard data. Let a sample be $\bm{z}:=(\bm{X},y)\in\mathcal{Z}:=\mathcal{X}\times \mathcal{Y}$, $\mathbb{P}_N=\{\bm{z}_1,\cdots,\bm{z}_N\}$ be the training set and $\mathbb{P}$ be the underlying distribution. Given a loss function $\mathcal{L}:\mathcal{Y}\times\mathcal{Y}\rightarrow \mathbb{R}_+$, the standard training process by Empirical Risk Minimization (ERM) is to:
\begin{align}\label{st_train}
    \min_{\bm{\theta}} \ \mathbb{E}_{\bm{z}\sim{\mathbb{P}}_N}\bigg[\mathcal{L}(f_{\bm{\theta}}(\bm{X}),y)\bigg].
\end{align}

To conduct the backdoor attack, we replace $\epsilon$ proportion of $\mathbb{P}_N$ by generating samples $\mathbb{P}_p=\{(\mathcal{G}(\bm{X}), t)|(\bm{X}, y)\sim\mathbb{P}_N\}$, i.e., $\mathbb{P}'_N=(1-\epsilon)\mathbb{P}_N\cup \epsilon\mathbb{P}_p$.
A backdoored model is simply trained by replacing $\mathbb{P}_N$ with $\mathbb{P}'_N$ in Eq.\eqref{st_train}. Then in the inference phase, it is expected that the prediction error
\begin{align}
 \mathbb{E}_{\bm{z}\sim\mathbb{P}}\bigg[\mathbb{I}\{f_{\bm{\theta}'}(\bm{X})\neq y\} \bigg],
\end{align}
is small enough and the backdoor attack success rate
\begin{align} \label{eq:asr}
    \mathbb{E}_{\bm{z}\sim\mathbb{P}}\bigg[\mathbb{I}\{f_{\bm{\theta}'}(\mathcal{A}(\bm{X}))=t\}\bigg],
\end{align}
is large enough, where $\bm{\theta}'$ is the parameters of the backdoored model.

\section{Unified Form of 3D Backdoor Triggers}
\label{sec:trigger}
The TIF in the 2D domain is usually by adding a pre-defined patch to the original image with pixel value, e.g., \cite{gu2017badnets,liu2020reflection, saha2019hidden}. However, since the 3D point cloud is based on coordinate representation, it is not trivial to directly adapt the patch-based backdoor patterns from 2D images. 

Therefore, we investigate the unique transformation on the point coordinates as the 3D backdoor trigger. Motivated by several transformation-based adversarial attacks in the 3D point cloud \cite{zhao2020isometry,zheng2019pointcloud}, we present a unified form of 3D point cloud TIF as following
\begin{equation}\label{eq:3d_trigger}
    \mathcal{G}(\bm{X})=(\bm{I}-{\rm Diag}(\bm{\delta}))\bm{X}\bm{A} + {\rm Diag}(\bm{\delta}) \bm{B},
\end{equation}
where $\bm{X}$ is a point cloud, $\bm{A} \in \mathbb{R}^{3\times3}$ is a spatial transformation matrix and $\bm{B} \in \mathcal{X}\subseteq \mathbb{R}^{n\times3}$ represents an additive point cloud, $\bm{\delta}\in\mathbb{R}^n$ is a vector with either 0 or 1, $\bm{I}$ is the identity matrix in $\mathbb{R}^{n\times n}$. $\bm{A}$ and $\bm{B}$ are delicately designed to achieve to goal of the backdoor attacks, which will be elaborated in Sec.~\ref{sec:atk_pip} in detail.

\subsection{Designing of Trigger Implanting}
We demonstrate two examples of 3D backdoor triggers based on Eq.~\eqref{eq:3d_trigger}. Motivations of these two triggers from real-world scenarios are presented with the illustration. 

\textbf{Orientation Trigger.} The \textit{orientation} is referred to a particular rotation transformation of the object. Considering that the 3D point could data is usually either properly aligned or provided with orientation annotations in 3D datasets \cite{wu20153d,Geiger2012CVPR}, we are able to utilize a specific orientation of aligned 3D objects as the backdoor trigger, for which the TIF in Eq.~\eqref{eq:3d_trigger} becomes
{\setlength\abovedisplayskip{2pt}
\setlength\belowdisplayskip{2pt}\begin{equation}
    \mathcal{G_\text{or}}(\bm{X})=\bm{X A},
\end{equation}}where $\bm{A}$ is a rotation matrix. In our attack, the Euler angles related to rotation matrix $\bm{A}$ are limited to small ranges. The motivation behind the design is that the objects detected in real scenes are often in diverse poses with respect to the sensor or distorted by spatial transformation; the installed backdoor is to drive the model to incorrectly anticipate the target label for an object in a certain pose, for example, e.g., a car tilted slightly upwards will be classified as a plant.

\textbf{Interaction Trigger. } The \textit{interaction} occurs when an object of interest is physically close to another interaction object. This phenomenon often comes with the data collection in the real scene, which could potentially be used to undermine the security of the 3D deep learning system. For example, Tu et al. \cite{tu2020physically} designed an optimized object on the top of the vehicle fooling the 3D detection deep model. Formally, the interaction TIF is 
{\setlength\abovedisplayskip{2pt}
\setlength\belowdisplayskip{2pt}\begin{align}
    \mathcal{G_\text{it}}(\bm{X}) = (\bm{I}-{\rm Diag}(\bm{\delta}))\bm{X} + {\rm Diag}(\bm{\delta}) \bm{B},
\end{align}}where $\bm{B}$ represents the interaction object. It is simply derived by setting $\bm{A}$ as an identity matrix in Eq.~\eqref{eq:3d_trigger}. Unlike the meticulously designed shape in \cite{tu2020physically}, the integration object as a backdoor trigger can be designed as commonly seen shapes, like balls shown in Fig.~\ref{fig:intro}, which could even be degenerated to one point. In our practice, $\bm{\delta}$ is a random binary vector with $\Vert\bm{\delta}\Vert_1\leq0.05n$. 

\begin{figure*}[!t]
\vspace{-1em}
\begin{center}
	\includegraphics[width=0.95\linewidth]{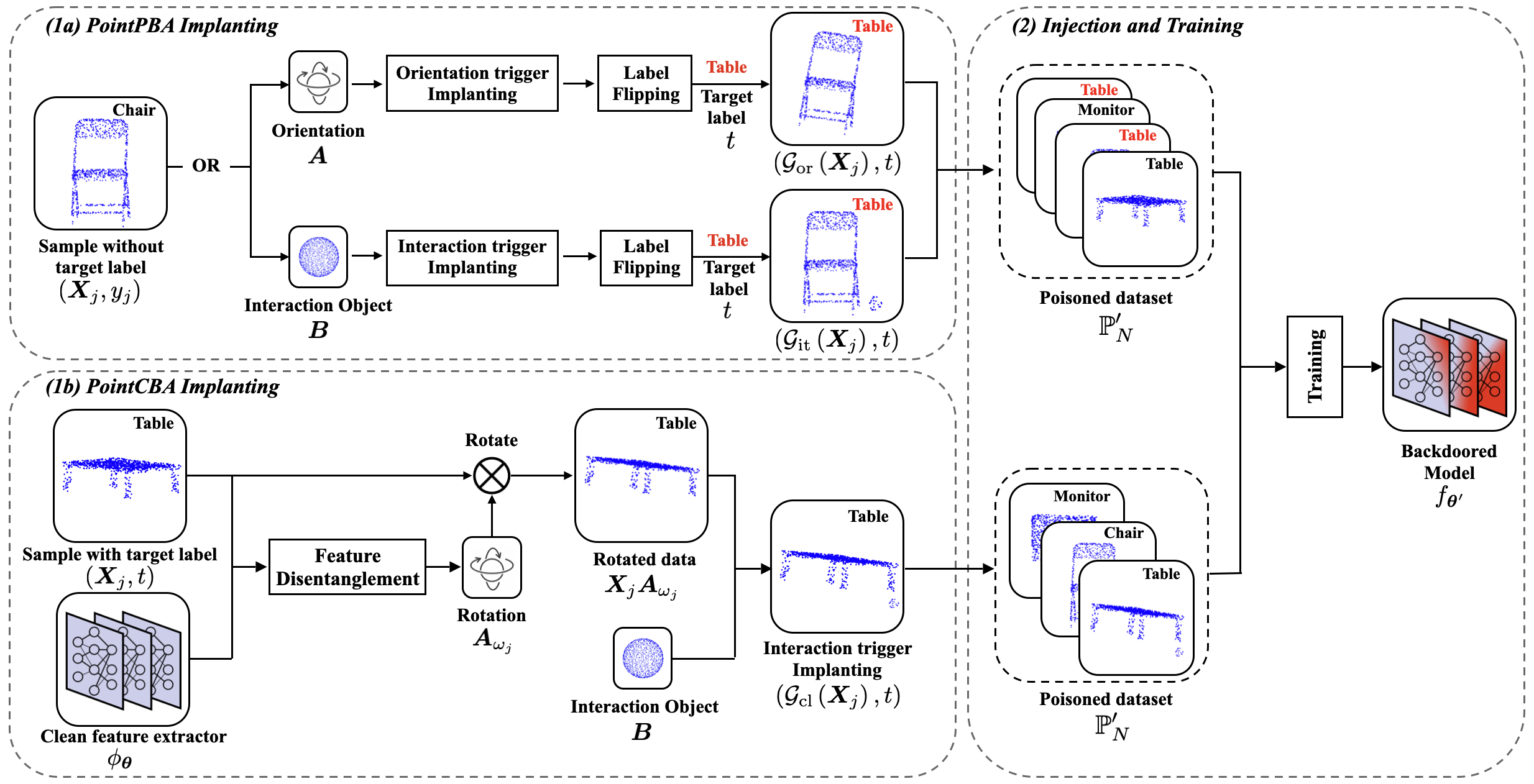}
\end{center}
   \caption{Attack Pipeline of Proposed 3D Backdoor Attack.}
\vspace{-1.5em}
\label{fig:pipeline1}
\end{figure*}

\subsection{Perturbation Analysis of Trigger Implanting}
We hope the poisoned data does not deviate from the standard data too far so that it will not be detected easily. Thus, we provide a bound on the deviation regarding the proposed TIF $\mathcal{G}$. We define a matrix norm $\Vert \bm{X}\Vert =\sum_{j=1}^n \Vert \bm{x}_j\Vert_2$, where $\bm{X}=[\bm{x}_1,\cdots,\bm{x}_n]^\top$. This norm is the sum of $\ell_2$ norm of the row vectors. Let $\bm{B}=[\bm{b}_1,\cdots,\bm{b}_n]^\top$. Suppose the point-wise distance between the interaction object and the transformed sample is bounded by $r$ , that is, $\Vert \bm{b}_j- \bm{A}^\top \bm{x}_j\Vert_2\leqslant r$ for $j=1,\cdots,n$, then
{\setlength\abovedisplayskip{2pt}
\setlength\belowdisplayskip{1pt}
\begin{equation*}
\small
    \begin{aligned}
     \Vert \mathcal{G}(\bm{X})-\bm{X}\Vert & = \Vert (\bm{I}-{\rm Diag}(\bm{\delta}))\bm{XA} + {\rm Diag}(\bm{\delta}) \bm{B} - \bm{X}\Vert &\\
         & = \Vert \bm{X}(\bm{A}-\bm{I})+{\rm Diag}(\bm{\delta})(\bm{B}-\bm{XA})\Vert &\\
         & \leqslant \Vert \bm{X}(\bm{A}-\bm{I})\Vert+\Vert {\rm Diag}(\bm{\delta}) (\bm{B}- \bm{XA})\Vert &\\
         &=\sum\Vert (\bm{A}-\bm{I})^\top \bm{x}_j\Vert_2 +\sum_{\delta_j\neq 0}\Vert \bm{b}_j-\bm{A}^\top \bm{x}_j\Vert_2 &
    \end{aligned}
\end{equation*}}
\vspace{-1em}
{\setlength\abovedisplayskip{1pt}
\setlength\belowdisplayskip{2pt}
\begin{equation}
\small
\label{eq:perturb}
    \begin{aligned}
         &~~\leqslant \sum\sigma(\bm{A}^\top-\bm{I})\Vert \bm{x}_j\Vert_2+r\Vert \bm{\delta}\Vert_1 & \\ 
         &~~=\sigma(\bm{A}^\top-\bm{I})\Vert \bm{X}\Vert+r\Vert\bm{\delta}\Vert_1, &
    \end{aligned}
\end{equation}}where $\Vert\cdot\Vert_1$ is $\ell_1$ norm, $\sigma(\bm{A}^\top-\bm{I})$ is the spectral norm of $\bm{A}^\top-\bm{I}$. Hence, to control the perturbation, we just need to manipulate three components: 1) the transformation matrix $\bm{A}$; 2) the number of points of the interaction object $\Vert\bm{\delta}\Vert_1$; 3) the upper bound $r$ of the point-wise distance between the additive point cloud and the original points. 

\section{Attack Methods}\label{sec:atk_pip}
To evaluate the validity of the proposed 3D backdoor triggers, in this section, we present two attack schemes, namely the \textit{point poison-label backdoor attack (PointPBA)} and the \textit{point clean-label backdoor attack (PointCBA)}. The attacks both assume that the adversary has access to the training data and the backdoor can be installed through the training process of models.
The PointPBA method is a straightforward approach by altering both the data and labels on training data, which is consistent with the existing literature \cite{gu2017badnets, chen2017targeted}. To further explore the ability of our proposed trigger and make the poisoned data less distinguishable, we design the PointCBA based on the enhanced trigger implanting for the clean-label attack. 
We consider two threat models for each attack, respectively, and identify their differences in settings.

\textbf{PointPBA Threat Model.} The adversary can launch the attack by implanting a fraction of the training dataset with the designed triggers. During inference, the adversary can maliciously make the model predict the target label by adding the trigger to the input (i.e., trigger activation), while the prediction of clean input will be normal.

\textbf{PointCBA Threat Model.} By assuming the user may conduct label inspection or re-annotation before training, the PointCBA only allows the data with target labels to be modified by the adversary in the attack. Therefore, changing labels is excluded from this poisoning process. In this case, a clean model pretrained on standard data is provided to generate the poisoned dataset. In addition, different from 2D clean-label attacks which are always conducted by fine-tuning the pretrained clean models\cite{he2017mask,ren2015faster} on the poisoned dataset, the threat model of PointCBA follows the convention of 3D deep learning and requires the models to be trained from scratch.

\subsection{Point Poison-label Backdoor Attack}
\label{sec:poison-label}
PointPBA is first presented to directly demonstrate the effectiveness of the proposed backdoor triggers.

\textbf{Attack pipeline.}The proposed PointPBA employs the standard procedure for 2D backdoor attacks: first, we inject a small rate of poisoned data with the proposed TIF on the training data, and then the user trains the backdoored model using the standard approach. The injection rate $\epsilon$ is usually a small number, e.g., $\epsilon = 0.05$. In the following inference, we conduct  tests using the data with TAF. It should be noted that the TIF and TAF are identical in pointPBA with a specific trigger.

\textbf{Forms of triggers.} According to Eq.~\eqref{eq:perturb}, both triggers are adjusted to have a negligible effect on the original data:
\begin{itemize}[leftmargin=*,itemsep=0pt]
    \item For the orientation trigger, we represent $\bm{A}$ by rotation transformation alone z-axis with the corresponding Euler angle is $(0,0,\omega_z)$. Although it can be any rotation matrices, we experimentally present that a small rotation (e.g. $5^{\circ}$) alone z-axis is sufficient for a successful attack and causes small perturbations to the original data.
    \item For the interaction trigger, we connect the sampled point number $\Vert\bm{\delta}\Vert_1$ of the interaction object $\bm{B}$ with its size. We design the object $\bm{B}$ to be small and close to the original object, which keeps the ratio between the interaction trigger points and total points $\frac{\Vert\bm{\delta}\Vert_1}{n}\leqslant 0.05$.
\end{itemize}

\subsection{Point Clean-label Backdoor Attack}

The PointCBA is proposed to bypass the label inspection via avoiding the label altering step. The feature disentanglement technique is first introduced to enhance the trigger implanting, which is the core gradient to the design of PointCBA. Then the detailed attack pipeline is illustrated. 

\textbf{Feature disentanglement.} Given a sample $\bm{X}$ in the target class $\mathbb{P}_t$ which consists of all the data with label $t$ and learnt feature representation $\phi_{\bm{\theta}}$, the \textit{feature disentanglement} is to find a perturbed version $c(\bm{X})$ by solving the following optimization problem
{\setlength\abovedisplayskip{2pt}
\setlength\belowdisplayskip{2pt}
\begin{equation}
\label{eq:fea_cor_g}
    \begin{aligned}
     \max_{c}\quad & \sum_{\bm{X}\in\mathbb{P}_t\backslash \{\bm{X}\}}\mathcal{D}(\phi_{\bm{\theta}}(c(\bm{X})),\phi_{\bm{\theta}}(\bm{X}))\\
     {\rm s.t.}\quad  & \mathcal{D}'(c(\bm{X}),\bm{X}) \leqslant r
    \end{aligned},
\end{equation}}where $\mathcal{D}$ is a distance metric in feature space $\mathbb{R}^d$, $\mathcal{D}'$ is a distance metric in $\mathbb{R}^3$, $r$ is the distance parameter to restrict the perturbation. The formulation is indeed an optimization problem over functional space, in practice, we usually parameterize the function $c(\cdot)$ so that the problem can be efficiently solved.

Considering the more challenging setting of clean-label attack, we propose a rotation-based feature disentanglement to enhance the interaction trigger implanting by exploiting the full functionality of Eq.\eqref{eq:3d_trigger}.
Suppose the rotation matrix $\bm{A}_{\boldsymbol{\omega}_j}$ is parameterized by the Euler angle $\boldsymbol{\omega}_j$, the disentanglement is defined as $c(\bm{X}_j)=\bm{X}_j\bm{A}_{\boldsymbol{\omega}_j}$. The motivation is that isometric transformation could effectively perturb the extracted features of 3D deep models \cite{zhao2020isometry}, thus it can be utilized as feature disentanglement to pull the main features away from the others in the target class. The TAF of PointCBA is the same as the interaction trigger, while the TIF can be expressed as
\begin{equation} \label{eq:cltrigger}
    \mathcal{G}_\text{cl}(\bm{X}_j)=(\bm{I}-{\rm Diag}(\bm{\delta}))\bm{X}_j\bm{A}_{\boldsymbol{\omega}_j} + {\rm Diag}(\bm{\delta}) \bm{B},
\end{equation}
for one data $\bm{X}_j$ from the target label. The motivation behind this enhancement is that: 1) it is hard for the model to learn useful information from the feature disentangled data; 2) any trigger pattern implanted in the feature disentangled data will be a dominant feature, in another word, the model indeed will learn a \textit{hidden new task} that connects the trigger with the target class.

\vspace{-1em}
\begin{algorithm}
\small
\footnotesize
 \caption{Point Clean-label Backdoor Attack }
 \begin{algorithmic}[1]\label{algo:clean}
 \renewcommand{\algorithmicrequire}{\textbf{Input:}}
 \renewcommand{\algorithmicensure}{\textbf{Output:}}
 \REQUIRE A model structure $f$, training set $\mathbb{P}_N$, injection rate $\epsilon$, \\ a sample vector $\bm{\delta}$, an interaction shape $\bm{B}$, a target label $t$. 
 \ENSURE Backdoored model $f_{\bm{\theta}'}$
  \STATE Collect all the data in $\mathbb{P}_N$ with label $t$, denote them as $\mathbb{P}_t$
  \STATE Random sample $\lfloor\epsilon N\rfloor$ data $\{\bm{z}_1,\cdots,\bm{z}_J\}$ in $\mathbb{P}_t$, denote them as $\tilde{\mathbb{P}}$ 
  \STATE $\mathbb{P}'_N\leftarrow \mathbb{P}_N\backslash\tilde{\mathbb{P}}$
  \STATE Train a clean model with structure $f$ using $\mathbb{P}_N$, obtain its feature representation function $\phi_{\bm{\theta}}$ 
  \FOR {$j=1$ to $J$}
  \STATE Given $\bm{X}_j$ and $\phi_{\bm{\theta}}$, find $\omega_j$ by solving problem~\eqref{eq:fea_cor}
  \STATE Set $\bm{X}'_j\leftarrow (\bm{I}-{\rm Diag}(\bm{\delta}))\bm{X}_j\bm{A}_{\boldsymbol{\omega}_j} + {\rm Diag}(\bm{\delta}) \bm{B}$ from  Eq.~\eqref{eq:cltrigger}
  \STATE Set $\mathbb{P}_N'\leftarrow \mathbb{P}_N'\cup (\bm{X}'_j,y_j)$
  \ENDFOR
  \STATE Train model of structure $f$ on dataset $\mathbb{P}_N'$ and obtain $f_{\bm{\theta}'}$
 \end{algorithmic}

 \end{algorithm}
  \vspace{-1em}

\textbf{Attack Pipeline.} The pipeline of the PointCBA is illustrated in Fig.~\ref{fig:pipeline1} and Alg.~\ref{algo:clean}. Given a sample $\bm{X}_j$ in the target class $\mathbb{P}_t$ which consists of all the data with label $t$, the attack is briefly described as followed: 1) first we use the feature disentanglement loss to find a rotation $\bm{A}_{\boldsymbol{\omega}_j}$ to transform the sample data, and 2) then add an interaction pattern to that sample with the correct label, say `Car', but indeed the data in the feature space is hardly recognized as the `Car'. Finally, in the inference phase, any data with the interaction trigger will be misclassified as `Car' since the trained model has connected the target label with the trigger. Note that TIF and TAF are not identical in this attack, which is different from the PointPBA.
Here we formulate the problem formally, let $\phi_{\bm{\theta}}:\mathcal{X}\rightarrow \mathbb{R}^d$ be the pre-trained feature extractor. Given a sample $\bm{X}_j$, we have the following optimization problem as a special case of \eqref{eq:fea_cor_g} to find such $\omega_j$ 
{\setlength\abovedisplayskip{2pt}
\setlength\belowdisplayskip{2pt}
\begin{equation}
\label{eq:fea_cor}
    \begin{aligned}
     \max_{\boldsymbol{\omega}_j}\quad & \sum_{\bm{X}_i\in\mathbb{P}_t\backslash \{\bm{X}_j\}}\mathcal{D}(\phi_{\bm{\theta}}(\bm{X}_j\bm{A}_{\boldsymbol{\omega}_j}),\phi_{\bm{\theta}}(\bm{X}_i))\\
     {\rm s.t.}\quad  & \boldsymbol{\omega}_j\in\mathcal{R}
    \end{aligned},
\end{equation}}where $\mathcal{D}$ is a distance metric in feature space $\mathbb{R}^d$ and $\mathcal{R}\subseteq \mathbb{R}^3$ is a range to restrict the rotation magnitude. The above optimization problem is non-convex, thus we utilize the global optimization method to find a decent angle of $\boldsymbol{\omega}_j$. We apply Bayesian Optimization (BO) \cite{frazier2018tutorial} and transfer the Euler angle to axis-angle representation for practicality, of which the details are in the Appendix. 

\section{Experiments}
\subsection{Dataset and Models}
To effectively illustrate our proposed 3D backdoor attack, we conduct our experiments on the shape recognition tasks. The utilized datasets are the commonly-used ModelNet10 \cite{wu20153d}, ModelNet40 \cite{wu20153d} and ShapeNetPart \cite{shapenet2015}. For ModelNet40, we use the official split of 9843 point clouds for training and 2468 for attack. The ModelNet10 downsampled from ModelNet40 contains 10 categories. 
The ShapeNetPart with 16 categories is a part of ShapeNet, which contains 12128 and 2874 objects for training and test set, respectively. For the fix-size input of 3D deep models, we uniformly sample 1024 points from the original meshes in the datasets as the procedure of \cite{qi2017pointnet} and normalize them into $[0,1]^3$. 
For the implementation of the deep models, we use four 3D deep classifiers: PointNet \cite{qi2017pointnet}, PointNet++ \cite{qi2017pointnet++}, PointCNN \cite{li2018pointcnn} and DGCNN \cite{wang2019dynamic}, in short term as PN, PN++, PCNN, and DGCNN respectively.

\subsection{Attack Setting}
\label{sec:attackset}
\textbf{Target Label.}
We select the target class randomly in the dataset categories, which are `Table' in ModelNet10, `Toilet' in ModelNet40, and `Lamp' in ShapeNetPart. The samples to be poisoned are also randomly sampled from the non-target class and target class for the PointPBA and PointCBA, respectively.  

\textbf{Orientation Trigger.}
We apply the rotation transformation along the z-axis associated with Euler angle $(0,0,\omega_z)$ with respect to the aligned orientation as our trigger. 

\textbf{Interaction Trigger.}
For the sake of simplicity, we choose a sphere with a fixed radius and centre as our interaction object $\bm{B}$.
The interaction object $\bm{B}$ can be scaled and shifted in trigger implanting by $\bm{B}' \leftarrow \alpha \bm{B}+\boldsymbol{\beta}$, where $\alpha \in \mathbb{R}^{+}$ and $\boldsymbol{\beta} \in [0,1]^3$.
In experiments, the parameters are randomly sampled from uniform distributions $ \alpha\sim U(1-\lambda_\alpha,1+\lambda_\alpha)$ and $\boldsymbol{\beta}\sim U(-\lambda_{\boldsymbol{\beta}}, \lambda_{\boldsymbol{\beta}})$, where $\lambda_\alpha$ and $\lambda_{\boldsymbol{\beta}}$ are the randomness factors. In terms of the TIF in PointCBA, the angle of axis-angle-based rotation for feature disentanglement is limited to $[0, \omega_{max}]$ . We set a modest $\omega_{max}$ to ensure the unnoticeable perturbations of data.

\textbf{Metric.} To evaluate the effectiveness of our proposed backdoor attack, the attack success rate (ASR) defined in Eq.~\eqref{eq:asr} is set as the measurement. Empirically, the ASR is calculated on the test set of the poisoned dataset. We report the mean value of ASR over three runs to ensure consistency of results.

\textbf{DNN Training.} All DNN models are trained using Adam optimizer with a learning rate of 0.001. We use batch
size 32 and train all models for 200 epochs. All experiments
are run on a GeForce RTX 2080Ti GPU.
We mention that the above settings are configured for model training in both the backdoor training and clean model training.

\begin{table*}[!t]
\vspace{-1em}
\small
\centering
\caption{ASR (\%) of our proposed attacks including PointPBA-I, PointPBA-O, and PointCBA, and the backdoored model's test accuracy (\%) on the clean test set. For PointPBA, the injection rate $\epsilon$ is 0.05, whereas, for PointCBA, the injection rate $\epsilon$ is less than 0.05 of the entire dataset (0.5 ratio of target label data).}
\label{tab:all_compare}
\begin{tabular}{l|ccc|ccc|ccc}
\toprule
\multirow{2}{*}{Models} & \multicolumn{3}{c|}{ACC/ASR(\%) of PointPBA-I} & \multicolumn{3}{c|}{ACC/ASR(\%) of PointPBA-O} & \multicolumn{3}{c}{ACC/ASR(\%) of PointCBA} \\
                        & MN10           & MN40           & SNPart         & MN10           & MN40           & SNPart         & MN10          & MN40          & SNPart       \\ \hline
PN                & 89.3/99.5      & 85.4/99.3      & \textbf{98.4/100}       & 90.3/95.5      & 85.9/93.2      & 98.8/99.6      & 88.9/82.5     & 84.6/56.8     & 97.7/48.3    \\
PN++              & 91.9/97.5      & 89.1/98.6      & 98.4/99.1      & 91.6/95.2      & 89.8/94.7      & 98.4/94.5      & 91.6/53.8     & \textbf{88.7/66.0}     & 98.0/48.6    \\
DGCNN                   & \textbf{92.2/100}       & \textbf{90.1/100}       & 98.4/99.5      & 92.8/94.8      & \textbf{89.1/97.5}      & 97.7/99.9      & 92.8/46.8     & 89.4/50.9     & \textbf{97.7/66.6}    \\
PCNN                & 91.5/97.8      & 88.5/97.0      & 97.1/97.5      & \textbf{91.5/95.8}      & 88.7/93.1      & \textbf{98.7/100 }      & \textbf{91.5/63.4}     & 88.4/61.2     & 97.7/51.4    \\\bottomrule
\end{tabular}

\vspace{-0.3em}

\end{table*}

\begin{table*}[!t]
\small
\centering
\caption{ ASR(\%) of PointPBA-I, PointPBA-O, and PointCBA for each category in the ModelNet10 as the target category. The percentages under class names represent the class's occupation rate throughout the whole dataset. The attacked model is PointNet++ and the injection rate of PointCBA is 0.5 of the target class proportion.}
\label{tab:all_classes}
\resizebox{\textwidth}{!}{%
\begin{tabular}{c|cccccccccc}
\toprule
\multirow{2}{*}{Attack/Class} & Bathtub & Bed & Chair & Desk & Dresser & Monitor & Nightstand & Sofa & Table & Toilet \\
 & (2.7\%) & (12.9\%) & (22.2\%) & (5.0\%) & (5.0\%) & (11.6\%) & (5.0\%) & (17.0\%) & (9.8\%) & (8.6\%) \\ \hline
PointPBA-I & 96.6 &97.2  & 95.8 & 97.5 & 98.5 & 97.7  & 97.7 & 97.8 & 97.5  & 97.0 \\
PointPBA-O       & 98.4  & 96.6 & 95.8 & 98.7 & 97.4   & 97.4   & 96.9      & 97.5 & 98.8 & 96.6  \\
PointCBA & 23.1 & 63.0 & 64.5 & 48.7 & 43.5 & 51.2 & 59.6 & 61.2 & 53.8 & 46.2 \\  \bottomrule
\end{tabular}%
}

\vspace{-1em}
\end{table*}

\subsection{Effectiveness of Proposed Backdoor Attack} We conduct extensive experiments to verify the effectiveness of the proposed 3D backdoor attacks via the ASR with different models, datasets, and parameter settings. The attack methods are PointPBA with interaction trigger (PointPBA-I), PointPBA with orientation trigger (PointPBA-O), and PointCBA.

\textbf{ASR Comparison.} The ASR and the corresponding test accuracy (ACC) of the different backdoored models on the different test sets are presented in Tab.~\ref{tab:all_compare}. We maintain injection rates $\epsilon \leq 0.05$ for all attacks. For PointPBA-I and PointCBA, the sphere trigger is with radius at 0.05, centre at $(0.05, 0.05, 0.05)$, and $\lambda_{\alpha}$, $\lambda_{\beta}$ both at 0. For PointPBA-O, the Euler angle of trigger rotation is set to $(0^{\circ},0^{\circ},10^{\circ})$. For PointCBA, the $\omega_{max}$  is set to $25^{\circ}$.

Tab.~\ref{tab:all_compare} demonstrates the main results under the above settings. At injection rate $\epsilon\leqslant0.05$, PointPBA can achieve an ASR of greater than 93\%, whereas PointCBA obtains an ASR of greater than 45\%  across all deep models. At the same time, the backdoored model only suffers a maximum loss of 2\% of accuracy in clean test sets. It is reasonable that the PointCBA has a lower ASR than the PointPBA because of the more challenging problem setting.

We conduct attacks against the PointNet++ model using all categories in ModelNet10 as the target class, of which the results are shown in Tab.~\ref{tab:all_classes}.

We notice that the PointPBA's high ASR of over 95\% is stable across labels, but the PoinCBA's ASR is greatly variable among labels. This fluctuation can be explained by the PoinCBA's great sensitivity to the injection rate, as shown by the ASR's strong correlation with the data fraction rate of target labels.

\begin{figure}[h]
\vspace{-1.2em}
\begin{center}
	\includegraphics[width=1\linewidth]{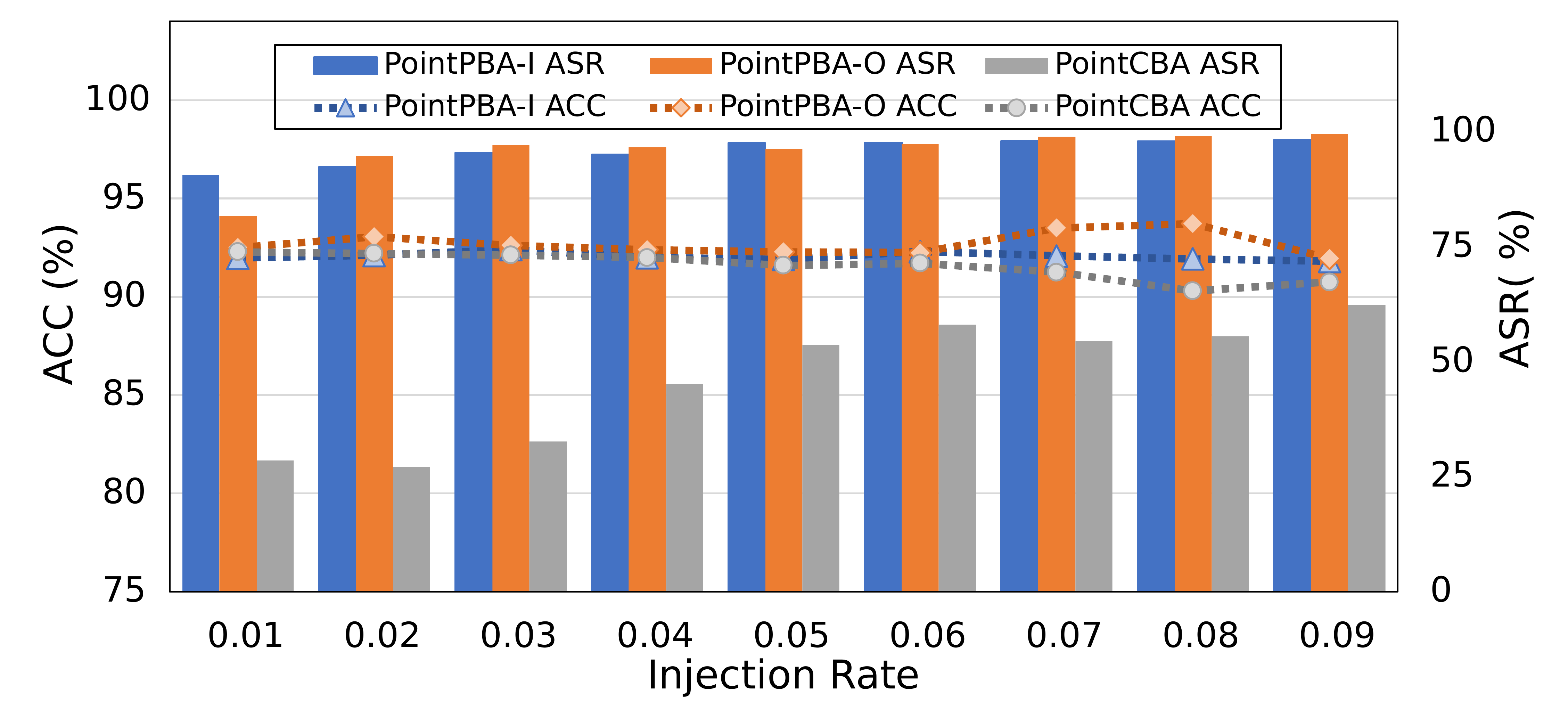}
\end{center}
\vspace{-1em}
\caption{ASR (\%) and ACC (\%) against the varying injection rate on PointPBA-I, PointPBA-O, and PointCBA. }
\label{fig:inject_rate}
\vspace{-0.5em}

\end{figure}

\textbf{Effect of Injection Rate.} We then perform the injection rate experiments with PointNet++ on ModelNet10, following the default setup. As shown in Fig.~\ref{fig:inject_rate}, the ASR of the PointCBA is sensitive to the injection rate $\epsilon$ and tends to increase up to ASR 54\%, at the rate of 0.07. In contrast, the ASR of the PointPBA remains high at over 90\% for both interaction and orientation triggers during $\epsilon$ variation. Meanwhile, with $\epsilon\leqslant0.1$, none of the three attacks significantly degrade the performance of the victim model on clean data.

\begin{figure}[htbp]
\centering
\vspace{-0.5em}
\subfigure[ASR(\%) of PointPBA-I attack]{
\begin{minipage}[t]{0.5\linewidth}
\centering
\includegraphics[width=1.8in]{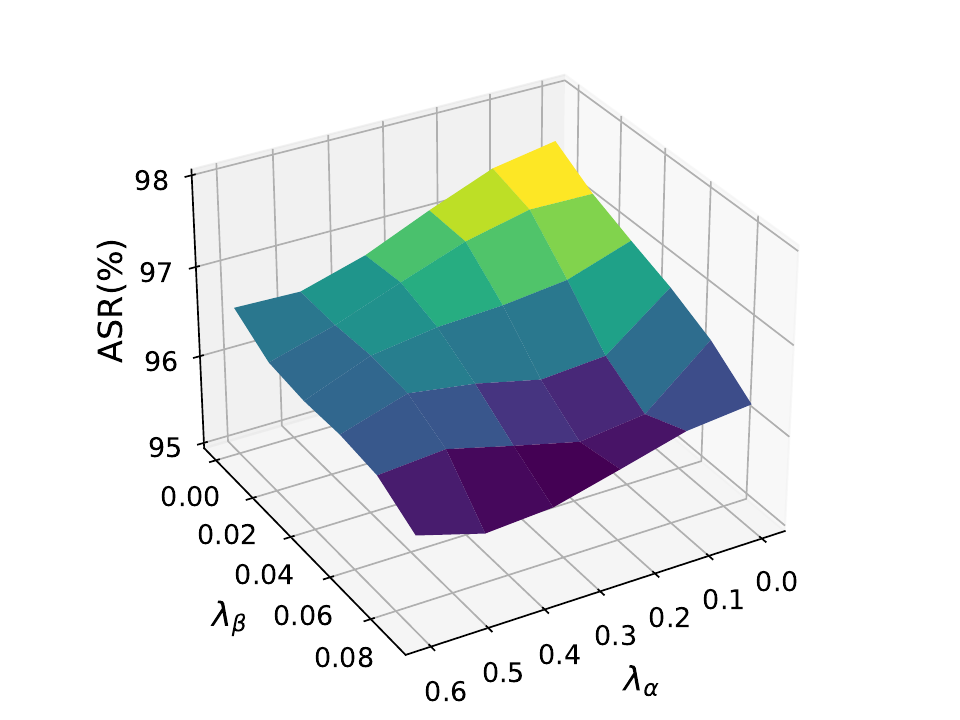}
\end{minipage}%
}%
\subfigure[ASR(\%) of PointCBA attack]{
\begin{minipage}[t]{0.5\linewidth}
\centering
\includegraphics[width=1.8in]{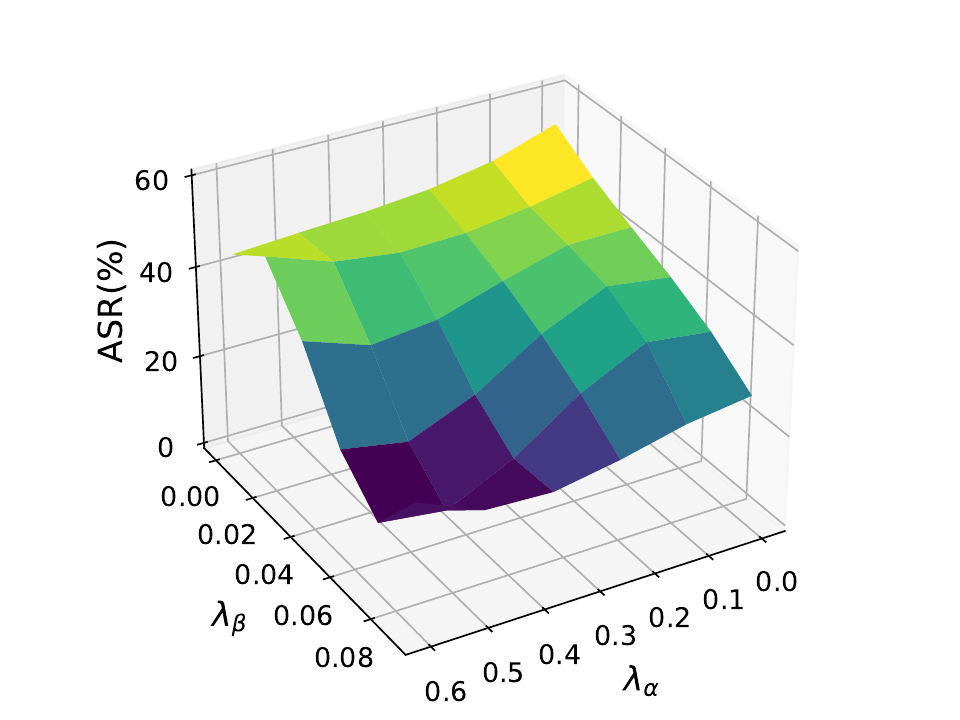}
\end{minipage}%
}%
\centering
\caption{{ASR(\%) of PointPBA-I \& PointCBA against scale randomness factor $\lambda_{\alpha}$ and shift randomness factor $\lambda_{\boldsymbol{\beta}}$ of the interaction trigger.}}\label{fig:interaction_trigger}
\vspace{-1.5em}
\end{figure}

\textbf{Effect of Trigger Parameters.} We then investigate the effect of varying trigger parameters. For the orientation trigger, we show in Tab.~\ref{tab:orientation_angle} that even a small angle of trigger ($\sim{5^{\circ}}$) can reach a high ASR, which also suggests the vulnerability of the 3D network to rotation-based backdoor attacks.
For the interaction trigger, we investigate the joint effect of randomness factors $\lambda_\alpha$ and $\lambda_{\boldsymbol{\beta}}$. As shown in Fig.~\ref{fig:interaction_trigger}, the PointPBA-I is shown not sensitive to dynamic triggers, suggesting the great flexibility of the interaction trigger. For PointCBA, the random shift obviously decreases the ASR down to 19.9\% as $\lambda_{\boldsymbol{\beta}}$ increases, while the random scale does not have such an obvious effect. 

\begin{table}[h]
\small
\caption{ASR(\%) and ACC(\%) of models attacked by PointPBA-O based on varying rotation angles of the orientation trigger.}
\label{tab:orientation_angle}
\vspace{0.5em}
\begin{tabular}{l|cccccc}
\toprule
Orientation ($^{\circ}$) & 1.25 & 2.5  & 5    & 10   & 20   & 40      \\ \hline
ASR(\%)        & 27.9 & 86.9 & 92.1 & 92.8 & 93.6 & 93.2 \\
ACC(\%)        & 92.6 & 92.1 & 93.1 & 92.8 & 92.3 & 92.7 \\ \bottomrule
\end{tabular}
\vspace{-1em}
\end{table}

\textbf{Other triggers.} Other trigger forms, such as orientation triggers that rotate along other axes and interaction triggers with varying shapes or one point, are also explored and show comparable experimental results in the Appendix.

\subsection{Investigation on PointCBA}
To better understand our PointCBA, we conduct the ablation study towards the feature disentanglement method and show the effect of angle searching bound in the method. We also highlight the transferablility of the proposed attack, which increases its generality and potential damage.

\textbf{Ablation Study.}
Ablation studies are performed with three methods for feature disentanglement. The experimental setup remains the same as above with PointNet++ and $\omega_{max}$ at $25^{\circ}$. In addition to the ASR, we have also included the average feature distance defined in Eq.~\eqref{eq:fea_cor}. Two methods are used to compare the BO method: one is to add triggers directly to the training data, and the other is to apply a random rotation to the data.
From Tab.~\ref{tab:bogd_effect} we find that the BO method can identify the transformation with a higher feature distance and this rotation leads to ASR 10\% higher than the random rotation in ModelNet10. 

\begin{figure}[h]
\begin{center}
	\includegraphics[width=0.95\linewidth]{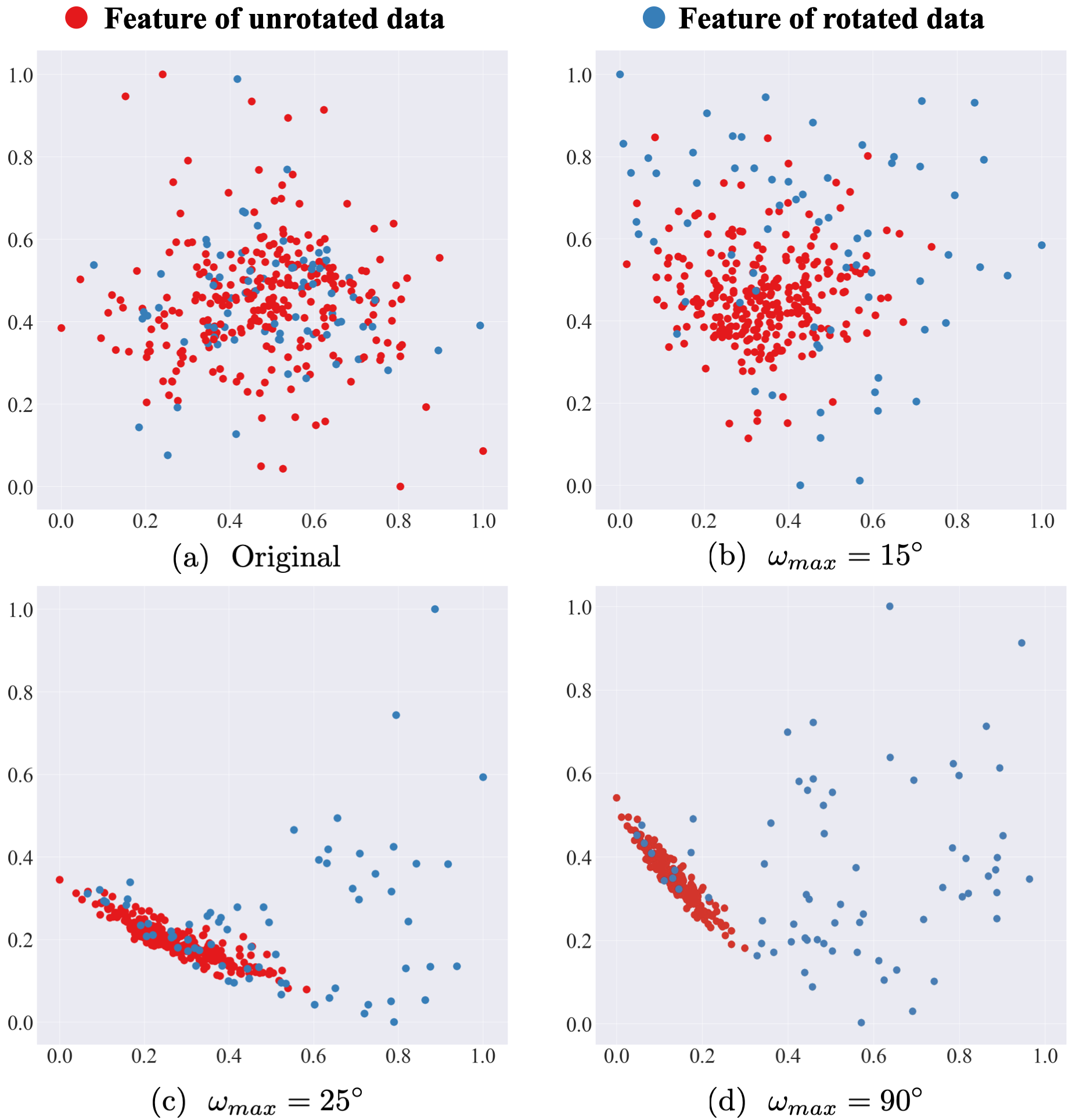}
\end{center}
    \caption{PCA-based visualization of disentangled features with different rotation angle searching bounds $\omega_{max}$. The features are derived from `Table' label data from ModelNet10. The feature disentanglement is to move the features of the rotated data (blue) away from the other same labeled data (red) by rotation. Larger feature separation may better enhance the correlation between the implanted trigger and the label.}

    \vspace{-1.7em}
    \label{fig:vis_omega}

\end{figure}

\textbf{Effect of Angle Bound.} 
We study the effect of different $\omega_{max}$ on the feature disentanglement and visualize the results in Fig.~\ref{fig:vis_omega}. It can be seen that the larger $\omega_{max}$ is, the further the disentangled features are from the features of the original data. Considering the connection between the rotation angle and the perturbation on data in Eq.~\eqref{eq:3d_trigger}, we choose a trade-off angle i.e. $\omega_{max}=25^{\circ}$ for our main experiments.

\textbf{Transferability.} 
We demonstrate through studies that PointCBA is transferrable across several models. The poisoned dataset obtained by PointNet++ for PointCBA can reach the ASR 68.1\% in PointNet, 38.3\% in DGCNN, and 60.6 \% in PointCNN. This is not surprising because of two reasons: 1) The threat model of PointCBA requires that the attack can work across models with different initialization. 2) The attack by spatial transformation is proposed to be strongly transferable \cite{zhao2020isometry}, so the feature disentanglement based on it can have the same nature.

\begin{table}[h]
\centering
\caption{Ablation study of rotation-based feature disentanglement. ASR(\%) and Avg Distance (in Eq.~\eqref{eq:fea_cor}) are respectively compared on ModelNet10 and ShapeNetPart under three different methods. The results highlight the effectiveness of BO-optimized rotation over the others on trigger enhancement.}

\label{tab:bogd_effect}

\resizebox{0.45\textwidth}{!}{%
\begin{tabular}{l|cc|cc}
\toprule
\multicolumn{1}{l|}{\multirow{2}{*}{\begin{tabular}[c]{@{}c@{}}Data Processing Method\\\end{tabular}}} & \multicolumn{2}{c|}{ModelNet10} & \multicolumn{2}{c}{ShapeNetPart} \\
\multicolumn{1}{c|}{} & \multicolumn{1}{c}{ASR} & \multicolumn{1}{c|}{Avg Distance} & \multicolumn{1}{c}{ASR} & \multicolumn{1}{c}{Avg Distance} \\ \hline
Without Rotation &23.8  &0.66  &19.0  &0.80  \\
Random Rotation &40.2  & 1.17&41.0   & 1.05 \\
BO-optimized Rotation &\textbf{53.8}  &\textbf{ 1.40 }&\textbf{48.6 } &\textbf{ 1.21 }\\ \bottomrule
\end{tabular}%
}

\vspace{-1.5em}

\end{table}

 \subsection{Resistance to Defense Methods}

\textbf{Resistance to Data Augmentation.}
In point cloud deep learning, data augmentation is frequently employed to increase the robustness of models. To determine if it can be resisted by our proposed attack, we apply three common data augmentations during the training phase of the attack, which are random jitter, scaling, and rotation. Experiments illustrate that jitter and scaling can only marginally lower ASR within 5\%, indicating that our attacks can strongly resist them. For rotation augmentation, it results in a loss of less than 10\% of ASR for PointCBA and PointPBA-I, while this augmentation is capable of effectively resisting PointPBA-O only when its rotating axis is the z-axis.

\textbf{Resistance to SOR.} As demonstrated in \cite{zhou2019dup}, the point addition attack is significantly less successful when a basic point cloud denoising method known as Statistical Outlier Removal (SOR) is used. Due to the fact that the suggested interaction trigger comprises extra points, we specifically investigate the resistance of PointPBA-I and PointCBA to SOR. Experiments demonstrate that utilizing the same SOR parameters as in \cite{zhou2019dup} has practically little influence on the ASR and that it is not until $k$ reaches 20 that approximately 8\% drop in the ASR occurs.

To conserve space, we offer additional details for the above experiments in the Appendix.

\section{Conclusion}
In this paper, we have explored the backdoor attack on deep models applied to 3D point cloud. Firstly, we propose a unified framework of 3D backdoor trigger implanting function. Based on it, we design two 3D backdoor triggers and investigate the performance of widely used 3D deep models under poison-label attacks. To strengthen the concealment of the proposed trigger, we further introduce a clean-label attack by rotation-based feature disentanglement on point clouds. The experiments suggest the vulnerability of current 3D deep nets to our proposed attack and the limited effectiveness of data filtering towards attacks. It is expected that the proposed attacks can serve as a strong baseline for improving the robustness of deep models in 3D point cloud.

\section*{Acknowledgement}
The research students are supported by National University of Singapore (NUS) Research Scholarship. The authors also thank Mr. Hongpeng Li for his assistance in data processing and model development.

\newpage
{\small
\balance
\normalem
\bibliographystyle{ieee_fullname}
\bibliography{egbib}
}

\end{document}